\documentclass[11pt,a4paper]{article}
\usepackage[hyperref]{naaclhlt2018}
\usepackage{times}
\usepackage{latexsym}
\usepackage{url}
\usepackage{amsmath}
\usepackage{amssymb}
\usepackage{graphicx}
\usepackage{booktabs}
\usepackage{tablefootnote}

\graphicspath{ {images/} }

\aclfinalcopy 

\newcommand{\squad}{\textsc{SQ\normalfont{u}AD}}
\newcommand{\lstm}{\textsc{LSTM}}
\newcommand{\bilstm}{\textsc{B\normalfont{i}LSTM}}
\newcommand{\rnn}{\textsc{RNN}}
\newcommand{\cnn}{\textsc{CNN}}

\newcommand{\rasor}{\textsc{R\normalfont{a}S\normalfont{o}R}}
\newcommand{\tr}{\textsc{TR}}
\newcommand{\trMlp}{\textsc{TR(MLP)}}

\newcommand{\rasorTrLstm}{\rasor{} + \tr{}}
\newcommand{\rasorTrMlp}{\rasor{} + \trMlp{}}

\newcommand{\trLmLi}{\tr{} + LM(L1)}
\newcommand{\trLmLii}{\tr{} + LM(L2)}

\newcommand{\rasorTrLstmLm}{\rasorTrLstm{} + \textsc{LM}}
\newcommand{\rasorTrLstmLmEmb}{\rasorTrLstmLm(emb)}
\newcommand{\rasorTrLstmLmLi}{\rasorTrLstmLm\textsc{(L1)}}
\newcommand{\rasorTrLstmLmLii}{\rasorTrLstmLm\textsc{(L2)}}

\title{Contextualized Word Representations for Reading Comprehension}

\author{Shimi Salant \\
  Tel-Aviv University \\
  {\tt {\small shimi.salant@cs.tau.ac.il}} \\\And
  Jonathan Berant \\
  Tel-Aviv University \\
  {\tt {\small joberant@cs.tau.ac.il}} \\}

\date{}

\begin{document}

\maketitle

\begin{abstract}
Reading a document and extracting an answer to a question about its content has attracted substantial attention recently. While most work has focused on the interaction between the question and the document, in this work we evaluate the importance of context when the question and document are processed independently. We take a standard neural architecture for this task, and show that by providing rich contextualized word representations from a large pre-trained language model as well as allowing the model to choose between context-dependent and context-independent word representations, we can obtain dramatic improvements and reach performance comparable to state-of-the-art on the competitive \textsc{SQuAD} dataset. 
\end{abstract}
\section{Introduction}
\label{sec:introduction}

Reading comprehension (RC) is a high-level task in natural language understanding that requires  reading a document and answering questions about its content. RC has attracted substantial attention over the last few years with the advent of large annotated datasets \cite{DBLP:journals/corr/HermannKGEKSB15,rajpurkar2016,DBLP:journals/corr/TrischlerWYHSBS16,DBLP:conf/nips/NguyenRSGTMD16,DBLP:conf/acl/JoshiCWZ17}, computing resources, and neural network models and optimization procedures \citep{DBLP:journals/corr/WestonBCM15,DBLP:conf/nips/SukhbaatarSWF15,DBLP:journals/corr/KumarISBEPOGS15}.

Reading comprehension models must invariably represent word tokens contextually, as a function of their encompassing sequence (document or question).
The vast majority of RC systems encode contextualized representations of words in both the document and question as hidden states of bidirectional RNNs \cite{hochreiter1997lstm,DBLP:journals/tsp/SchusterP97,DBLP:conf/ssst/ChoMBB14}, and focus model design and capacity around question-document \textit{interaction},
carrying out calculations where information from both is available \citep{seo2017,xiong2017,fusionnet,DBLP:conf/acl/WangYWCZ17}.

Analysis of current RC models has shown that models tend to react to simple word-matching between the question and document \citep{DBLP:conf/emnlp/JiaL17}, as well as benefit from explicitly providing matching information in model inputs \citep{DBLP:journals/corr/HuPQ17,DBLP:conf/acl/ChenFWB17,DBLP:conf/conll/WeissenbornWS17}. In this work, we hypothesize that the still-relatively-small size of RC datasets drives this behavior, which leads to models that make limited use of context when representing word tokens.

To illustrate this idea, we take a model that carries out only basic question-document interaction and prepend to it a module that produces token embeddings by explicitly gating between contextual and non-contextual representations (for both the document and question). This simple addition already places the model's performance on par with recent work, and allows us to demonstrate the importance of context.

Motivated by these findings, we turn to a semi-supervised setting in which we leverage a language model, pre-trained on large amounts of data, as a sequence encoder which forcibly facilitates context utilization. We find that model performance substantially improves, reaching accuracy comparable to state-of-the-art on the competitive \squad{} dataset, showing that contextual word representations captured by the language model are beneficial for reading comprehension.
\footnote{Our complete code base is available at {\small{\url{http://github.com/shimisalant/CWR}}}.}

\section{Contextualized Word Representations}
\label{sec:model}

\paragraph{Problem definition} We consider the task of \textit{extractive} reading comprehension: given a paragraph of text
$\mathbf{p}=(p_1,\dots,p_n)$
and a question
$\mathbf{q}=(q_1,\dots,q_m)$, an answer span $(p_l,\dots,p_r)$ is to be extracted, i.e., a pair of indices $1\leq l \leq r \leq n$ into $\mathbf{p}$ are to be predicted.

When encoding a word token in its encompassing sequence (question or passage), we are interested in allowing extra computation over the sequence and evaluating the extent to which context is utilized in the resultant representation.
To that end, we employ a \textit{re-embedding} component in which a contextual and a non-contextual representation are explicitly combined per token. Specifically, for a sequence of word-embeddings $w_1,\dots,w_k$ with $w_t\in\mathbb{R}^{d_w}$, the re-embedding of the $t$-th token $w_t'$ is the result of a Highway layer \cite{srivastava2015} and is defined as:
\begin{align*}
w_t' &= g_t \odot w_t + (1-g_t) \odot z_t \\
g_t &= \sigma(W_g x_t + U_g u_t) \\
z_t &= tanh(W_z x_t + U_z u_t)
\end{align*}
where $x_t$ is a function strictly of the word-type of the $t$-th token, $u_t$ is a function of the enclosing sequence, $W_g,W_z,U_g,Uz$ are parameter matrices, and $\odot$ the element-wise product operator. We set $x_t=[w_t;c_t]$, a concatenation of $w_t$ with $c_t\in\mathbb{R}^{d_c}$ where the latter is a character-based representation of the token's word-type produced via a \cnn{} over character embeddings \cite{kim2014}. We note that word-embeddings $w_t$ are pre-trained \citep{pennington2014} and are kept fixed during training, as is commonly done in order to reduce model capacity and mitigate overfitting. We next describe different formulations for the contextual term $u_t$.

\paragraph{RNN-based token re-embedding \small{(TR)}}
Here we set $\{u_1,\dots,u_k\}=\bilstm{}(x_1,\dots,x_k)$ as the hidden states of the top layer in a stacked \bilstm{} of multiple layers, each uni-directional \lstm{} in each layer having $d_h$ cells and $u_k\in\mathbb{R}^{2d_h}$.

\paragraph{LM-augmented token re-embedding \small{(TR+LM)}}
The simple module specified above allows better exploitation of the context that a token appears in, if such exploitation is needed and is not learned by the rest of the network, which operates over $w_1',\dots,w_k'$. 
Our findings in Section \ref{sec:evaluation_and_analysis} indicate that context is crucial but that in our setting it may be utilized to a limited extent.

We hypothesize that the main determining factor in this behavior is the relatively small size of the data and its distribution, which does not require using long-range context in most examples. Therefore, we leverage a strong language model that was pre-trained on large corpora as a fixed encoder which supplies additional contextualized token representations. We denote these representations as $\{o_1,\dots,o_k\}$ and set $u_t=[u_t';o_t]$ for $\{u_1',\dots,u_k'\}=\textsc{B\normalfont{i}LSTM}(x_1,\dots,x_k)$.

The LM we use is from \citet{jozefowicz2016},\footnote{Named \textsc{BIG LSTM+CNN INPUTS} in that work and available at \small{\url{http://github.com/tensorflow/models/tree/master/research/lm_1b}}.} trained on the One Billion Words Benchmark dataset \cite{DBLP:journals/corr/ChelbaMSGBK13}. It consists of an initial layer which produces character-based word representations, followed by two stacked \lstm{} layers and a softmax prediction layer. The hidden state outputs of each \lstm{} layer are projected down to a lower dimension via a bottleneck layer \cite{sak2014}. We set $\{o_1,\dots,o_k\}$ to either the projections of the first layer, referred to as \trLmLi{}, or those of the second one, referred to as \trLmLii{}.

With both re-embedding schemes, we use the resulting representations $w_1',\dots,w_k'$ as a drop-in replacement for the word-embedding inputs fed to a standard model, described next.

\section{Base model}
\label{sec:base_model}

We build upon \citet{lee2016}, who proposed the \rasor{} model. For word-embedding inputs $q_1,\dots,q_m$ and $p_1,\dots,p_n$ of dimension ${d_w}$, \rasor{} consists of the following components:

\paragraph{Passage-independent question representation}
The question is encoded via a BiLSTM $\{v_1,\dots,v_m\}=\textsc{B\normalfont{i}LSTM}(q_1,\dots,q_m)$ and the resulting hidden states are summarized via attention \citep{bahdanau2014,DBLP:conf/emnlp/ParikhT0U16}: $q^{indep}=\sum_{j=1}^{m} \alpha_{j} v_j\in\mathbb{R}^{2d_h}$.
The attention coefficients $\mathbf{\alpha}$ are normalized logits \{$\alpha_1,\dots,\alpha_m\}=\textsc{\normalfont{softmax}}(s_1,\dots,s_m)$ where $s_j=w_q^T \cdot \textsc{FF}(v_j)$ for a parameter vector $w_q\in\mathbb{R}^{d_f}$ and $\textsc{FF}(\cdot)$ a single layer feed-forward network.

\paragraph{Passage-aligned question representations}
For each passage position $i$, the question is encoded via attention operated over its word-embeddings $q_{i}^{align}=\sum_{j=1}^{m}\beta_{ij} q_j\in\mathbb{R}^{d_w}$.
The coefficients $\beta_i$ are produced by normalizing the logits $\{s_{i1},\dots,s_{im}\}$, where $s_{ij}=\textsc{FF}(q_j)^T \cdot \textsc{FF}(p_i)$.

\paragraph{Augmented passage token representations}
Each passage word-embedding $p_i$ is concatenated with its corresponding $q_i^{align}$ and with the independent $q^{indep}$ to produce $p_i^*=[p_i;q_i^{align};q^{indep}]$, and a BiLSTM is operated over the resulting vectors: $\{h_1,\dots,h_n\}=\textsc{B\normalfont{i}LSTM}(p_1^*,\dots,p_n^*)$.

\paragraph{Span representations} A candidate answer span $a=(l,r)$ with $l \leq r$ is represented as the concatenation of the corresponding augmented passage representations: $h_a^* = [h_l;h_r]$. In order to avoid quadratic runtime, only spans up to length 30 are considered.

\paragraph{Prediction layer}
Finally, each span representation $h_a^*$ is transformed to a logit $s_{a} = w_c^T \cdot \textsc{FF}(h_a^*)$ for a parameter vector $w_c\in\mathbb{R}^{d_f}$, and these logits are normalized to produce a distribution over spans. Learning is performed by maximizing the log-likelihood of the correct answer span.

\section{Evaluation and Analysis}
\label{sec:evaluation_and_analysis}

We evaluate our contextualization scheme on the \squad{} dataset \cite{rajpurkar2016} which consists of 100,000+  paragraph-question-answer examples, crowdsourced from Wikipedia articles.

\paragraph{Importance of context}
We are interested in evaluating the effect of our \rnn{}-based re-embedding scheme on the performance of the downstream base model. However, the addition of the re-embedding module incurs additional depth and capacity for the resultant model. We therefore compare this model, termed \rasorTrLstm{}, to a setting in which re-embedding is non-contextual, referred to as \rasorTrMlp{}. Here we set $u_t=\textsc{MLP}(x_t)$, a multi-layered perceptron on $x_t$, allowing for the additional computation to be carried out on word-level representations without any context and matching the model size and hyper-parameter search budget of \rasorTrLstm{}.
In Table \ref{table:dev_results} we compare these two variants over the development set and observe superior performance by the contextual one, illustrating the benefit of contextualization and specifically per-sequence contextualization which is done separately for the question and for the passage.

\begin{figure}[t!]
\vspace{5 pt}
\includegraphics[scale=0.6]{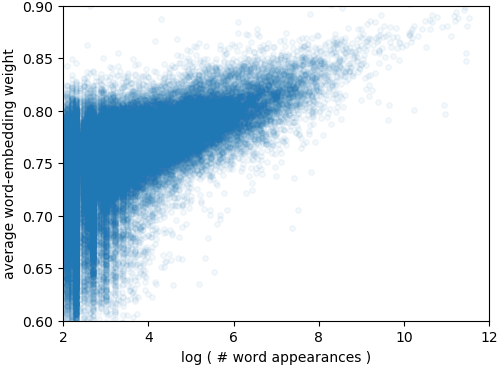}
\caption{Average gate activations.}
\label{fig:reembedder_gates}
\vspace{5 pt}
\end{figure}

\paragraph{Context complements rare words} Our formulation lends itself to an inspection of the different dynamic weightings computed by the model for interpolating between contextual and non-contextual terms. In Figure \ref{fig:reembedder_gates} we plot the average gate value $g_t$ for each word-type, where the average is taken across entries of the gate vector and across all occurrences of the word in both passages and questions. This inspection reveals the following:
On average, the less frequent a word-type is, the smaller are its gate activations, i.e., the re-embedded representation of a rare word places less weight on its fixed word-embedding and more on its contextual representation, compared to a common word. This  highlights a problem with maintaining fixed word representations: albeit pre-trained on extremely large corpora, the embeddings of rare words need to be complemented with information emanating from their context. Our specific parameterization allows observing this directly, but it may very well be an implicit burden placed on any contextualizing encoder such as a vanilla \bilstm.

\begin{table}[t]
\centering
\footnotesize
\begin{tabular}{l c c}
\toprule
Model & EM & F1 \\
\midrule
\text{\rasor{} (base model)} & 70.6 & 78.7 \\
\midrule
\text{\rasorTrMlp} & 72.5 & 79.9 \\
\text{\rasorTrLstm} & 75.0 & 82.5 \\
\midrule
\text{\rasorTrLstmLmEmb} & 75.8 & 83.0 \\
\text{\rasorTrLstmLmLi} & \textbf{77.0} & \textbf{84.0} \\
\text{\rasorTrLstmLmLii} & 76.1 & 83.3 \\
\bottomrule
\end{tabular}
\caption{
Results on SQuAD's development set. The EM metric measures an exact-match between a predicted answer and a correct one and the F1 metric measures the overlap between their bag of words.
}
\label{table:dev_results}
\end{table}

\begin{table}[t]
\centering
\footnotesize
\begin{tabular}{l c c}
\toprule
Model & EM & F1 \\
\midrule
BiDAF + Self Attention + ELMo {[}1{]} & 78.6 & 85.8 \\
\text{\rasorTrLstmLmLi} {[}2{]} & 77.6  & 84.2  \\
SAN {[}3{]} & 76.8 & 84.4 \\
r-net {[}4{]} & 76.5 & 84.3 \\
FusionNet {[}5{]} & 76.0 & 83.9 \\
Interactive AoA Reader+ {[}6{]} & 75.8 & 83.8 \\
\text{\rasorTrLstm} {[}7{]} & 75.8 & 83.3 \\
DCN+ {[}8{]} & 75.1 & 83.1 \\
Conductor-net {[}9{]} & 73.2 & 81.9 \\
$\cdots$ & & \\
\text{\rasor{} (base model)} {[}10{]} & 70.8 & 78.7 \\
\bottomrule
\end{tabular}
\caption[caption for test table]{
\setcounter{footnote}{2}
Single-model results on \squad{}'s test set.\footnotemark \\
{[}1{]} \citet{DBLP:journals/corr/abs-1802-05365}
{[}2,7{]} This work.
{[}3{]} \citet{san2017}
{[}4{]} \citet{DBLP:conf/acl/WangYWCZ17}
{[}5{]} \citet{fusionnet}
{[}6{]} \citet{DBLP:conf/acl/CuiCWWLH17}
{[}8{]} \citet{DBLP:journals/corr/abs-1711-00106}
{[}9{]} \citet{DBLP:journals/corr/abs-1710-10504}
{[}10{]} \citet{lee2016}
}
\label{tab:test_results}
\end{table}

\paragraph{Incorporating language model representations}
Supplementing the calculation of token re-embeddings with the hidden states of a strong language model proves to be highly effective. In Table \ref{table:dev_results} we list development set results for using either the LM hidden states of the first stacked \lstm{} layer or those of the second one. We additionally evaluate the incorporation of that model's word-type representations (referred to as \rasorTrLstmLmEmb{}), which are based on character-level embeddings and are naturally unaffected by context around a word-token.

Overall, we observe a significant improvement with all three configurations, effectively showing the benefit of training a QA model in a semi-supervised fashion \cite{DBLP:conf/nips/DaiL15} with a large language model.
Besides a crosscutting boost in results, we note that the performance due to utilizing the LM hidden states of the first \lstm{} layer significantly surpasses the other two variants. This may be due to context being most strongly represented in those hidden states as the representations of LM(emb) are non-contextual by definition and those of LM(L2) were optimized (during LM training) to be similar to parameter vectors that correspond to word-types and not to word-tokens.

In Table \ref{tab:test_results} we list the top-scoring single-model published results on \squad{}'s test set, where we observe \rasorTrLstmLmLi{} ranks second in EM, despite having only minimal question-passage interaction which is a core component of other works.
An additional evaluation we carry out is following \citet{DBLP:conf/emnlp/JiaL17}, which demonstrated the proneness of current QA models to be fooled by distracting sentences added to the paragraph.
In Table \ref{table:adv_results} we list the single-model results reported thus far and observe that the utilization of LM-based representations carried out by \rasorTrLstmLmLi{} results in improved robustness to adversarial examples.

\begin{table}[t]
\centering
\footnotesize
\begin{tabular}{l c c}
\toprule
Model & AddSent & AddOneSent \\
\midrule
\text{\rasorTrLstmLmLi} {[}1{]} &  47.0 & 57.0 \\
Mnemonic Reader {[}2{]} & 46.6 & 56.0 \\
\text{\rasorTrLstm} {[}3{]} & 44.5 & 53.9 \\
MPCM {[}4{]} & 40.3 & 50.0 \\
\text{\rasor{} (base model)} {[}5{]} & 39.5 & 49.5 \\
ReasoNet {[}6{]} & 39.4 & 50.3 \\
jNet {[}7{]} & 37.9 & 47.0 \\
\bottomrule
\end{tabular}
\caption{
\phantom{x} Single-model F1 on adversarial \squad{}. \\
{[}1,3{]} This work.
{[}2{]} \citet{DBLP:journals/corr/HuPQ17}
{[}4{]} \citet{DBLP:journals/corr/WangMHF16}
{[}5{]} \citet{lee2016}
{[}6{]} \citet{DBLP:conf/kdd/ShenHGC17}
{[}7{]} \citet{DBLP:journals/corr/ZhangZCDWJ17}
}
\label{table:adv_results}
\end{table}

\footnotetext{From \squad{}'s leaderboard per Dec 13, 2017. \small{\url{http://rajpurkar.github.io/SQuAD-explorer}}}

\section{Experimental setup}
\label{sec:experimental_setup}

We use pre-trained GloVe embeddings \cite{pennington2014} of dimension $d_w=300$ and produce character-based word representations via $d_c=100$ convolutional filters over character embeddings as in \citet{seo2017}.
For all \textsc{B\normalfont{i}LSTM}s, hyper-parameter search included the following values, with model selection being done according to validation set results (underlined): number of stacked \textsc{B\normalfont{i}LSTM} layers $(1,\underline{2},3)$, number of cells $d_h$ $(50,100,\underline{200})$, dropout rate over input $(0.4, 0.5, \underline{0.6})$, dropout rate over hidden state $(0.05, \underline{0.1}, 0.15)$. 
To further regularize models, we  employed word dropout \citep{DBLP:conf/acl/IyyerMBD15,DBLP:conf/nips/DaiL15} at rate $(0.05, 0.1, \underline{0.15})$ and couple \lstm{} input and forget gate as in \citet{greff2016}.
All feed-forward networks and the \textsc{MLP} employed the ReLU non-linearity \cite{nair2010} with dropout rate $(\underline{0.2},0.3)$, where the single hidden layer of the \textsc{FF}s was of dimension $d_f=(50,\underline{100})$ and the best performing \textsc{MLP} consisted of 3 hidden layers of dimensions $865$, $865$ and $400$.
For optimization, we used Adam \cite{kingma2015} with batch size $80$.

\section{Related Work}

Our use of a Highway layer with \rnn{}s is related to Highway \lstm{} \cite{DBLP:conf/icassp/ZhangCYYKG16} and Residual \lstm{} \cite{DBLP:journals/corr/KimEL17}. The goal in those works is to effectively train many stacked \lstm{} layers and so highway and residual connections are introduced 
into the definition of the  \lstm{} function. 
Our formulation is external to that definition, with the specific goal of gating between \lstm{} hidden states and fixed word-embeddings.

Multiple works have shown the efficacy of semi-supervision for NLP tasks \cite{anders2013}. Pre-training a LM in order to initialize the weights of an encoder has been reported to improve generalization and training stability for sequence classification \citep{DBLP:conf/nips/DaiL15} as well as translation and summarization \citep{DBLP:conf/emnlp/RamachandranLL17}.

Similar to our work, \citet{DBLP:journals/corr/PetersABP17} utilize the same pre-trained LM from \citet{jozefowicz2016} for sequence tagging tasks, keeping encoder weights fixed during training. Their formulation includes a backward LM and uses the hidden states from the top-most stacked \lstm{} layer of the LMs, whereas we also consider reading the hidden states of the bottom one, which substantially improves performance.
In parallel to our work, \citet{DBLP:journals/corr/abs-1802-05365} have successfully leveraged pre-trained LMs for several tasks, including RC, by utilizing representations from all layers of the pre-trained LM.

In a transfer-learning setting, \citet{DBLP:conf/nips/McCannBXS17} pre-train an attentional encoder-decoder model for machine translation and show improvements across a range of tasks when incorporating the hidden states of the encoder as additional fixed inputs for downstream task training.

\section{Conclusion}

In this work we examine the importance of context for the task of reading comprehension. We present a neural module that gates contextual and non-contextual representations and observe gains due to context utilization. Consequently, we  inject contextual information into our model by integrating a pre-trained language model through our suggested module and find that it substantially improves results, reaching state-of-the-art performance on the \squad{} dataset.

\section*{Acknowledgements}
We thank the anonymous reviewers for their constructive comments. This work was supported by the Israel Science Foundation, grant 942/16, and by the Yandex Initiative in Machine Learning.

\bibliography{naaclhlt2018}

\begin{thebibliography}{}
\expandafter\ifx\csname natexlab\endcsname\relax\def\natexlab#1{#1}\fi

\bibitem[{Bahdanau et~al.(2015)Bahdanau, Cho, and Bengio}]{bahdanau2014}
Dzmitry Bahdanau, Kyunghyun Cho, and Yoshua Bengio. 2015.
\newblock Neural machine translation by jointly learning to align and
  translate.
\newblock In {\em ICLR\/}.

\bibitem[{Chelba et~al.(2013)Chelba, Mikolov, Schuster, Ge, Brants, and
  Koehn}]{DBLP:journals/corr/ChelbaMSGBK13}
Ciprian Chelba, Tomas Mikolov, Mike Schuster, Qi~Ge, Thorsten Brants, and
  Phillipp Koehn. 2013.
\newblock One billion word benchmark for measuring progress in statistical
  language modeling.
\newblock {\em CoRR\/} abs/1312.3005.

\bibitem[{Chen et~al.(2017)Chen, Fisch, Weston, and
  Bordes}]{DBLP:conf/acl/ChenFWB17}
Danqi Chen, Adam Fisch, Jason Weston, and Antoine Bordes. 2017.
\newblock Reading wikipedia to answer open-domain questions.
\newblock In {\em ACL\/}.

\bibitem[{Cho et~al.(2014)Cho, van Merrienboer, Bahdanau, and
  Bengio}]{DBLP:conf/ssst/ChoMBB14}
Kyunghyun Cho, Bart van Merrienboer, Dzmitry Bahdanau, and Yoshua Bengio. 2014.
\newblock On the properties of neural machine translation: Encoder-decoder
  approaches.
\newblock In {\em EMNLP\/}.

\bibitem[{Cui et~al.(2017)Cui, Chen, Wei, Wang, Liu, and
  Hu}]{DBLP:conf/acl/CuiCWWLH17}
Yiming Cui, Zhipeng Chen, Si~Wei, Shijin Wang, Ting Liu, and Guoping Hu. 2017.
\newblock Attention-over-attention neural networks for reading comprehension.
\newblock In {\em ACL\/}.

\bibitem[{Dai and Le(2015)}]{DBLP:conf/nips/DaiL15}
Andrew~M. Dai and Quoc~V. Le. 2015.
\newblock Semi-supervised sequence learning.
\newblock In {\em NIPS\/}.

\bibitem[{Greff et~al.(2016)Greff, Srivastava, Koutn{\' i}k, Steunebrink, and
  Schmidhuber}]{greff2016}
Klaus Greff, Rupesh~Kumar Srivastava, Jan Koutn{\' i}k, Bas~R. Steunebrink, and
  Jurgen Schmidhuber. 2016.
\newblock Lstm: A search space odyssey.
\newblock In {\em IEEE Transactions on Neural Networks and Learning Systems\/}.

\bibitem[{Hermann et~al.(2015)Hermann, Kocisk{\'{y}}, Grefenstette, Espeholt,
  Kay, Suleyman, and Blunsom}]{DBLP:journals/corr/HermannKGEKSB15}
Karl~Moritz Hermann, Tom{\'{a}}s Kocisk{\'{y}}, Edward Grefenstette, Lasse
  Espeholt, Will Kay, Mustafa Suleyman, and Phil Blunsom. 2015.
\newblock Teaching machines to read and comprehend.
\newblock {\em CoRR\/} abs/1506.03340.

\bibitem[{Hochreiter and Schmidhuber(1997)}]{hochreiter1997lstm}
S.~Hochreiter and J.~Schmidhuber. 1997.
\newblock Long short-term memory.
\newblock {\em Neural Computation\/} 9(8):1735--1780.

\bibitem[{Hu et~al.(2017)Hu, Peng, and Qiu}]{DBLP:journals/corr/HuPQ17}
Minghao Hu, Yuxing Peng, and Xipeng Qiu. 2017.
\newblock Mnemonic reader for machine comprehension.
\newblock {\em CoRR\/} abs/1705.02798.

\bibitem[{Huang et~al.(2017)Huang, Zhu, Shen, and Chen}]{fusionnet}
Hsin{-}Yuan Huang, Chenguang Zhu, Yelong Shen, and Weizhu Chen. 2017.
\newblock Fusionnet: Fusing via fully-aware attention with application to
  machine comprehension.
\newblock {\em CoRR\/} abs/1711.07341.

\bibitem[{Iyyer et~al.(2015)Iyyer, Manjunatha, Boyd{-}Graber, and
  III}]{DBLP:conf/acl/IyyerMBD15}
Mohit Iyyer, Varun Manjunatha, Jordan~L. Boyd{-}Graber, and Hal~Daum{\'{e}}
  III. 2015.
\newblock Deep unordered composition rivals syntactic methods for text
  classification.
\newblock In {\em ACL\/}.

\bibitem[{Jia and Liang(2017)}]{DBLP:conf/emnlp/JiaL17}
Robin Jia and Percy Liang. 2017.
\newblock Adversarial examples for evaluating reading comprehension systems.
\newblock In {\em EMNLP\/}. pages 2011--2021.

\bibitem[{Joshi et~al.(2017)Joshi, Choi, Weld, and
  Zettlemoyer}]{DBLP:conf/acl/JoshiCWZ17}
Mandar Joshi, Eunsol Choi, Daniel~S. Weld, and Luke Zettlemoyer. 2017.
\newblock Triviaqa: {A} large scale distantly supervised challenge dataset for
  reading comprehension.
\newblock In {\em ACL\/}.

\bibitem[{J{\'{o}}zefowicz et~al.(2016)J{\'{o}}zefowicz, Vinyals, Schuster,
  Shazeer, and Wu}]{jozefowicz2016}
Rafal J{\'{o}}zefowicz, Oriol Vinyals, Mike Schuster, Noam Shazeer, and Yonghui
  Wu. 2016.
\newblock Exploring the limits of language modeling.
\newblock {\em CoRR\/} abs/1602.02410.

\bibitem[{Kim et~al.(2017)Kim, El{-}Khamy, and
  Lee}]{DBLP:journals/corr/KimEL17}
Jaeyoung Kim, Mostafa El{-}Khamy, and Jungwon Lee. 2017.
\newblock Residual {LSTM:} design of a deep recurrent architecture for distant
  speech recognition.
\newblock {\em CoRR\/} abs/1701.03360.

\bibitem[{Kim(2014)}]{kim2014}
Yoon Kim. 2014.
\newblock Convolutional neural networks for sentence classification.
\newblock In {\em EMNLP\/}.

\bibitem[{Kingma and Ba(2015)}]{kingma2015}
Diederik~P. Kingma and Jimmy~Lei Ba. 2015.
\newblock Adam: A method for stochastic optimization.
\newblock In {\em ICLR\/}.

\bibitem[{Kumar et~al.(2015)Kumar, Irsoy, Su, Bradbury, English, Pierce,
  Ondruska, Gulrajani, and Socher}]{DBLP:journals/corr/KumarISBEPOGS15}
Ankit Kumar, Ozan Irsoy, Jonathan Su, James Bradbury, Robert English, Brian
  Pierce, Peter Ondruska, Ishaan Gulrajani, and Richard Socher. 2015.
\newblock Ask me anything: Dynamic memory networks for natural language
  processing.
\newblock {\em CoRR\/} abs/1506.07285.

\bibitem[{Lee et~al.(2016)Lee, Salant, Kwiatkowski, Parikh, Das, and
  Berant}]{lee2016}
Kenton Lee, Shimi Salant, Tom Kwiatkowski, Ankur~P. Parikh, Dipanjan Das, and
  Jonathan Berant. 2016.
\newblock Learning recurrent span representations for extractive question
  answering.
\newblock {\em CoRR\/} abs/1611.01436.

\bibitem[{Liu et~al.(2017{\natexlab{a}})Liu, Wei, Mao, and
  Chikina}]{DBLP:journals/corr/abs-1710-10504}
Rui Liu, Wei Wei, Weiguang Mao, and Maria Chikina. 2017{\natexlab{a}}.
\newblock Phase conductor on multi-layered attentions for machine
  comprehension.
\newblock {\em CoRR\/} abs/1710.10504.

\bibitem[{Liu et~al.(2017{\natexlab{b}})Liu, Shen, Duh, and Gao}]{san2017}
Xiaodong Liu, Yelong Shen, Kevin Duh, and Jianfeng Gao. 2017{\natexlab{b}}.
\newblock Stochastic answer networks for machine reading comprehension.
\newblock {\em CoRR\/} abs/1712.03556.

\bibitem[{McCann et~al.(2017)McCann, Bradbury, Xiong, and
  Socher}]{DBLP:conf/nips/McCannBXS17}
Bryan McCann, James Bradbury, Caiming Xiong, and Richard Socher. 2017.
\newblock Learned in translation: Contextualized word vectors.
\newblock In {\em NIPS\/}.

\bibitem[{Nair and Hinton(2010)}]{nair2010}
Vinod Nair and Geoffrey~E. Hinton. 2010.
\newblock Rectified linear units improve restricted boltzmann machines.
\newblock In {\em ICML\/}.

\bibitem[{Nguyen et~al.(2016)Nguyen, Rosenberg, Song, Gao, Tiwary, Majumder,
  and Deng}]{DBLP:conf/nips/NguyenRSGTMD16}
Tri Nguyen, Mir Rosenberg, Xia Song, Jianfeng Gao, Saurabh Tiwary, Rangan
  Majumder, and Li~Deng. 2016.
\newblock {MS} {MARCO:} {A} human generated machine reading comprehension
  dataset.
\newblock In {\em NIPS Workshop on Cognitive Computation\/}.

\bibitem[{Parikh et~al.(2016)Parikh, T{\"{a}}ckstr{\"{o}}m, Das, and
  Uszkoreit}]{DBLP:conf/emnlp/ParikhT0U16}
Ankur~P. Parikh, Oscar T{\"{a}}ckstr{\"{o}}m, Dipanjan Das, and Jakob
  Uszkoreit. 2016.
\newblock A decomposable attention model for natural language inference.
\newblock In {\em EMNLP\/}.

\bibitem[{Pennington et~al.(2014)Pennington, Socher, and
  Manning}]{pennington2014}
Jeffrey Pennington, Richard Socher, and Christopher~D. Manning. 2014.
\newblock Glove: Global vectors for word representation.
\newblock In {\em {EMNLP}\/}.

\bibitem[{Peters et~al.(2017)Peters, Ammar, Bhagavatula, and
  Power}]{DBLP:journals/corr/PetersABP17}
Matthew~E. Peters, Waleed Ammar, Chandra Bhagavatula, and Russell Power. 2017.
\newblock Semi-supervised sequence tagging with bidirectional language models.
\newblock {\em CoRR\/} abs/1705.00108.

\bibitem[{Peters et~al.(2018)Peters, Neumann, Iyyer, Gardner, Clark, Lee, and
  Zettlemoyer}]{DBLP:journals/corr/abs-1802-05365}
Matthew~E. Peters, Mark Neumann, Mohit Iyyer, Matt Gardner, Christopher Clark,
  Kenton Lee, and Luke Zettlemoyer. 2018.
\newblock Deep contextualized word representations.
\newblock {\em CoRR\/} abs/1802.05365.

\bibitem[{Rajpurkar et~al.(2016)Rajpurkar, Zhang, Lopyrev, and
  Liang}]{rajpurkar2016}
Pranav Rajpurkar, Jian Zhang, Konstantin Lopyrev, and Percy Liang. 2016.
\newblock Squad: 100, 000+ questions for machine comprehension of text.
\newblock In {\em EMNLP\/}.

\bibitem[{Ramachandran et~al.(2017)Ramachandran, Liu, and
  Le}]{DBLP:conf/emnlp/RamachandranLL17}
Prajit Ramachandran, Peter~J. Liu, and Quoc~V. Le. 2017.
\newblock Unsupervised pretraining for sequence to sequence learning.
\newblock In {\em EMNLP\/}.

\bibitem[{Sak et~al.(2014)Sak, Senior, and Beaufays}]{sak2014}
Hasim Sak, Andrew~W. Senior, and Fran{\c{c}}oise Beaufays. 2014.
\newblock Long short-term memory recurrent neural network architectures for
  large scale acoustic modeling.
\newblock In {\em INTERSPEECH\/}.

\bibitem[{Schuster and Paliwal(1997)}]{DBLP:journals/tsp/SchusterP97}
Mike Schuster and Kuldip~K. Paliwal. 1997.
\newblock Bidirectional recurrent neural networks.
\newblock {\em {IEEE} Trans. Signal Processing\/} 45(11):2673--2681.

\bibitem[{Seo et~al.(2016)Seo, Kembhavi, Farhadi, and Hajishirzi}]{seo2017}
Min~Joon Seo, Aniruddha Kembhavi, Ali Farhadi, and Hannaneh Hajishirzi. 2016.
\newblock Bidirectional attention flow for machine comprehension.
\newblock {\em CoRR\/} abs/1611.01603.

\bibitem[{Shen et~al.(2017)Shen, Huang, Gao, and
  Chen}]{DBLP:conf/kdd/ShenHGC17}
Yelong Shen, Po{-}Sen Huang, Jianfeng Gao, and Weizhu Chen. 2017.
\newblock Reasonet: Learning to stop reading in machine comprehension.
\newblock In {\em SIGKDD\/}.

\bibitem[{S{\o}gaard(2013)}]{anders2013}
Anders S{\o}gaard. 2013.
\newblock Semi-supervised learning and domain adaptation for nlp .

\bibitem[{Srivastava et~al.(2015)Srivastava, Greff, and
  Schmidhuber}]{srivastava2015}
Rupesh~Kumar Srivastava, Klaus Greff, and Jürgen Schmidhuber. 2015.
\newblock Highway networks.
\newblock In {\em ICML 2015 Deep Learning workshop\/}.

\bibitem[{Sukhbaatar et~al.(2015)Sukhbaatar, Szlam, Weston, and
  Fergus}]{DBLP:conf/nips/SukhbaatarSWF15}
Sainbayar Sukhbaatar, Arthur Szlam, Jason Weston, and Rob Fergus. 2015.
\newblock End-to-end memory networks.
\newblock In {\em NIPS\/}.

\bibitem[{Trischler et~al.(2016)Trischler, Wang, Yuan, Harris, Sordoni,
  Bachman, and Suleman}]{DBLP:journals/corr/TrischlerWYHSBS16}
Adam Trischler, Tong Wang, Xingdi Yuan, Justin Harris, Alessandro Sordoni,
  Philip Bachman, and Kaheer Suleman. 2016.
\newblock Newsqa: {A} machine comprehension dataset.
\newblock {\em CoRR\/} abs/1611.09830.

\bibitem[{Wang et~al.(2017)Wang, Yang, Wei, Chang, and
  Zhou}]{DBLP:conf/acl/WangYWCZ17}
Wenhui Wang, Nan Yang, Furu Wei, Baobao Chang, and Ming Zhou. 2017.
\newblock Gated self-matching networks for reading comprehension and question
  answering.
\newblock In {\em ACL\/}.

\bibitem[{Wang et~al.(2016)Wang, Mi, Hamza, and
  Florian}]{DBLP:journals/corr/WangMHF16}
Zhiguo Wang, Haitao Mi, Wael Hamza, and Radu Florian. 2016.
\newblock Multi-perspective context matching for machine comprehension.
\newblock {\em CoRR\/} abs/1612.04211.

\bibitem[{Weissenborn et~al.(2017)Weissenborn, Wiese, and
  Seiffe}]{DBLP:conf/conll/WeissenbornWS17}
Dirk Weissenborn, Georg Wiese, and Laura Seiffe. 2017.
\newblock Making neural {QA} as simple as possible but not simpler.
\newblock In {\em CoNLL\/}.

\bibitem[{Weston et~al.(2015)Weston, Bordes, Chopra, and
  Mikolov}]{DBLP:journals/corr/WestonBCM15}
Jason Weston, Antoine Bordes, Sumit Chopra, and Tomas Mikolov. 2015.
\newblock Towards ai-complete question answering: {A} set of prerequisite toy
  tasks.
\newblock {\em CoRR\/} abs/1502.05698.

\bibitem[{Xiong et~al.(2017{\natexlab{a}})Xiong, Zhong, and
  Socher}]{DBLP:journals/corr/abs-1711-00106}
Caiming Xiong, Victor Zhong, and Richard Socher. 2017{\natexlab{a}}.
\newblock {DCN+:} mixed objective and deep residual coattention for question
  answering.
\newblock {\em CoRR\/} abs/1711.00106.

\bibitem[{Xiong et~al.(2017{\natexlab{b}})Xiong, Zhong, and Socher}]{xiong2017}
Caiming Xiong, Victor Zhong, and Richard Socher. 2017{\natexlab{b}}.
\newblock Dynamic coattention networks for question answering.
\newblock In {\em ICLR\/}.

\bibitem[{Zhang et~al.(2017)Zhang, Zhu, Chen, Dai, Wei, and
  Jiang}]{DBLP:journals/corr/ZhangZCDWJ17}
Junbei Zhang, Xiao{-}Dan Zhu, Qian Chen, Li{-}Rong Dai, Si~Wei, and Hui Jiang.
  2017.
\newblock Exploring question understanding and adaptation in
  neural-network-based question answering.
\newblock {\em CoRR\/} abs/1703.04617.

\bibitem[{Zhang et~al.(2016)Zhang, Chen, Yu, Yao, Khudanpur, and
  Glass}]{DBLP:conf/icassp/ZhangCYYKG16}
Yu~Zhang, Guoguo Chen, Dong Yu, Kaisheng Yao, Sanjeev Khudanpur, and James~R.
  Glass. 2016.
\newblock Highway long short-term memory {RNNS} for distant speech recognition.
\newblock In {\em ICASSP\/}.

\end{thebibliography}
\bibliographystyle{acl_natbib}

\end{document}